\documentclass[10pt,twocolumn,letterpaper]{article}

\usepackage[pagenumbers]{cvpr} 

\definecolor{cvprblue}{rgb}{0.21,0.49,0.74}
\usepackage[pagebackref,breaklinks,colorlinks,allcolors=cvprblue]{hyperref}

\def\paperID{15557} 
\def\confName{CVPR}
\def\confYear{2025}

\title{VMix: Improving Text-to-Image Diffusion Model \\ with
Cross-Attention Mixing Control}

\author{
\textbf{Shaojin Wu$^{1}$,
Fei Ding$^1$\thanks{Corresponding author.},
Mengqi Huang$^{1,2}$,
Wei Liu$^1$,
Qian He$^1$}
\smallskip 
\\
$^1$ByteDance Inc, $^2$University of Science and Technology of China
\smallskip 
\\
\tt\small\{wushaojin, dingfei.212, liuwei.jikun, heqian\}@bytedance.com 
\tt\small\{huangmq\}@mail.ustc.edu.cn
}

\begin{document}
\maketitle
\begin{abstract}
While diffusion models show extraordinary talents in text-to-image generation, they may still fail to generate highly aesthetic images. More specifically, there is still a gap between the generated images and the real-world aesthetic images in finer-grained dimensions including color, lighting, composition, \emph{etc.} In this paper, we propose Cross-Attention \textbf{V}alue \textbf{Mix}ing Control (\textbf{VMix}) Adapter, a plug-and-play aesthetics adapter, to upgrade the quality of generated images while maintaining generality across visual concepts by (1) disentangling the input text prompt into the content description and aesthetic description by the initialization of aesthetic embedding, and (2) integrating aesthetic conditions into the denoising process through value-mixed cross-attention, with the network connected by zero-initialized linear layers. Our key insight is to enhance the aesthetic presentation of existing diffusion models by designing a superior condition control method, all while preserving the image-text alignment. Through our meticulous design, VMix is flexible enough to be applied to community models for better visual performance without retraining. To validate the effectiveness of our method, we conducted extensive experiments, showing that VMix outperforms other state-of-the-art methods and is compatible with other community modules (\emph{e.g.}, LoRA, ControlNet, and IPAdapter) for image generation. The project page is \url{https://vmix-diffusion.github.io/VMix/}.
\end{abstract}    
\section{Introduction}
\label{sec:intro}
\begin{figure}[h]
\centering
\includegraphics[scale=0.48]{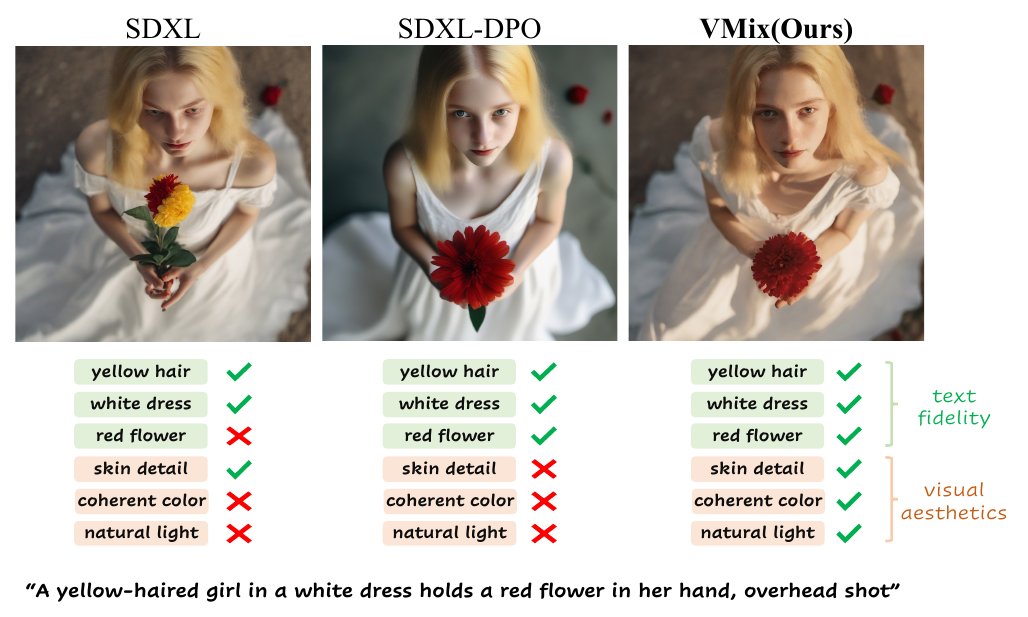}
    \caption{Comparison of text fidelity and visual aesthetics between SDXL~\cite{podell2023sdxl}, DPO~\cite{wallace2024diffusion}, and our VMix. DPO can generate attributes that SDXL fails to produce, but it fails to align with human visual fine-grained preferences. Our method achieves better text fidelity and visual aesthetics simultaneously.}
    \label{Figure 0}
\end{figure}
The past few years have witnessed the flourish in the field of text-to-image generation, especially the advent of large-scale pretrained text-to-image diffusion models\cite{rombach2022high,nichol2021glide,ramesh2022hierarchical,feng2023ernie,podell2023sdxl}, allowing human to create visually compelling images using text prompts conveniently.
Although these large models can produce overall high-quality images in visual realism and textual alignment, the current results still exhibit significant gaps from human expectations in various aspects, such as generated images with unnatural lighting, distorted human bodies, or supersaturated colors.
These misalignment with human expectation is fatal for the real-world applications of AI-generated content such as film production, since human beings are highly sensitive to these \emph{``devils in the details"}.
Therefore, the key challenge has become how to accurately align the generated images with human preference across various aspects.

Existing works have made considerable efforts to improve image quality to meet human preferences, which can be primarily categorized into two streams.
The first stream focuses on fine-tuning the pre-trained text-to-image models either based on an exceptionally high-quality sub-dataset\cite{dai2023emu}, or based on reward models via reinforcement learning\cite{liang2024rich, xu2024imagereward, kirstain2023pick} and direct preference optimization(DPO)\cite{wallace2024diffusion}.
The latter stream\cite{si2023freeu, he2024freestyle} instead focuses on investigating the generation behavior of pre-trained diffusion models itself to improve its generation stability.
For example, FreeU\cite{si2023freeu} proposed re-weighting the contributions sourced from skip connections and backbones in the denoising model to strengthen the denoising ability while simultaneously enhancing details.
In summary, existing methods align the generation results with human preference by improving the overall generation quality in terms of visual realism or textual consistency.

However, in this study, we argue that existing methods fail to align \emph{fine-grained human preference} for visually generated content.
Images favored by human beings should excel across various fine-grained aesthetic dimensions simultaneously, such as natural light, coherent color, and reasonable composition.
On the one hand, these fine-grained aesthetic demands can not be simply addressed by augmenting detailed textual descriptions for pretrained diffusion models to understand.
The reason is that their text encoders (\emph{e.g.}, CLIP\cite{radford2021learning} or T5\cite{raffel2020exploring}) are primarily trained for capturing high-level semantics and lacking the accurate awareness for these ineffable visual aesthetics.
On the other hand, the optimization direction for overall image generation quality is neither equivalent to nor consistent with the direction for these fine-grained aesthetic dimensions. 
For instance, while overall better-generated results may exhibit greater textual alignment, they might suffer from poorer visual composition, as depicted in \cref{Figure 0}.

To address this challenge, we introduce \textbf{VMix}, a novel plug-and-play adapter designed to systematically bridge the aesthetic quality gap between generated images and real-world counterparts across various aesthetic dimensions. 
We finetune the adapter on a hand-selected subset of exceptionally high-quality images derived from a large corpus. Inspired by universal photography standards, which encompass aspects like color, lighting, composition, and focus~\cite{dai2023emu}, we label these images across various aesthetic dimensions.
During training, we freeze the base model and employ the LoRA\cite{hu2021lora} method to ensure practical applicability.
We further design two specialized modules to incorporate these aesthetic labels as additional conditions into the U-Net~\cite{ronneberger2015u} architecture. The first, termed the \emph{aesthetic embedding initialization module}, pre-processes the aesthetic textual data, initializing it into embeddings that align with the corresponding images. 
This step is essential only at the commencement of training. 
Once training begins, we map the aesthetic labels of various images to embeddings by referencing the initial results. 
To better integrate this embedding into the U-Net, we introduce the second module, the \emph{cross-attention mixing control module}, which aims to minimize adverse effects on image-text alignment without directly altering the attention maps.
Our extensive experiments demonstrate that VMix can be seamlessly integrated with various base models, significantly enhancing their aesthetic performance. Moreover, VMix exhibits excellent compatibility with community modules (i.e., ControlNet\cite{zhang2023adding}, IP-Adapter\cite{ye2023ip}, and LoRA\cite{hu2021lora}), thereby providing the community with greater creative capabilities.

In summary, our main contributions are:
\begin{itemize}
    \item We analyze and explore the differences in generated images across fine-grained aesthetic dimensions, proposing the disentanglement of these attributes in the text prompt to provide a clear direction for model optimization. 
    \item We introduce VMix, which disentangles the input text prompt into content description and aesthetic description, offering improved guidance to the model via a novel condition control method called value-mixed cross-attention.
    \item The proposed VMix approach is universally effective for existing diffusion models, serving as a plug-and-play aesthetic adapter that is highly compatible with community modules.
\end{itemize}

\section{Related Work}
\label{sec:Related Work}

\subsection{Text-to-Image Models} 
The method of generating images from given textual descriptions has been extensively explored. GAN-based works have demonstrated impressive capabilities in producing realistic images~\cite{tao2022df,xu2018attngan,zhang2021cross}. Generative transformer methods usually train large-scale autoregressive transformers in a discrete token space to model the generation process~\cite{ramesh2021zero,yu2022scaling,yu2023scaling}. More recently, diffusion models have been applied in text-to-image tasks, achieving state-of-the-art results in generating high-fidelity images~\cite{rombach2022high,podell2023sdxl,esser2024scaling}. Given a text prompt, the model typically converts the text into a latent vector with the help of a pretrained language model~\cite{radford2021learning}, and then generates images from pure Gaussian noise through a iterative denoising process. Although these models have demonstrated remarkable capabilities in image generation, they struggle to ensure that the generated images perform well across multiple aesthetic dimensions.

\subsection{Improving Text-to-Image Models} 
Despite the significant breakthroughs in text-to-image diffusion models, numerous challenges persist when it comes to simultaneously ensuring text fidelity and visual aesthetics. Researchers are approaching these challenges from various angles to find solutions. Emu~\cite{dai2023emu} highlights the importance of high-quality data, demonstrating that models fine-tuned on such data can achieve further improvements in their generated results. FreeU~\cite{si2023freeu} enhances image generation quality by adjusting the connection weights at different levels of the U-Net, without requiring additional training. DPO~\cite{wallace2024diffusion} optimizes the denoising process by generating images that are more aligned with human preferences compared to less favored images. Unlike these approaches, we propose a new conditional control method that aligns with human aesthetics across fine-grained dimensions while retaining the original model's semantic understanding capabilities.

\begin{figure*}[ht]
	\centerline{\includegraphics[scale=0.46]{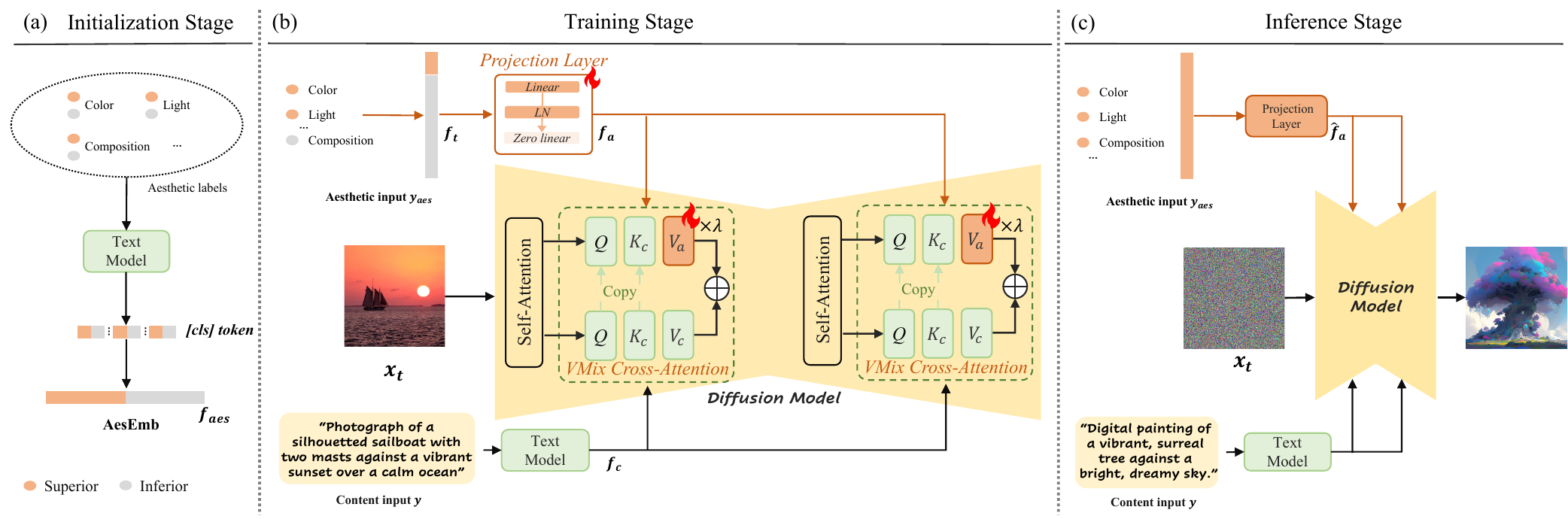}}
	\caption{Illustration of of VMix. (a)In the initialization stage, pre-defined aesthetic labels are transformed into [CLS] tokens through CLIP, thereby obtaining AesEmb, which only need to be processed once at the beginning of training. (b)In the training stage, a project layer first maps the input aesthetic description $y_{aes}$ into an embedding $f_a$ of the same token dimension as the content text embedding $f_t$. The text embedding $f_t$ is then integrated into the denoising network through value-mixed cross-attention. (c)In the inference stage, VMix extract all positive aesthetic embedding from AesEmb to form the aesthetic input, along with the content input, is fed into the model for the denoising process.}
	\label{Figure 1}
\end{figure*}

\subsection{Controlling Text-to-Image Models} 
Text-to-image models can generate results that match specific tasks or personalized content or styles by incorporating additional controlling conditions~\cite{mou2023t2i,wei2023elite}. Diverse conditional control approaches typically vary in the specific conditional features they introduce or the distinct points at which these conditions are injected into the process. ControlNet~\cite{zhang2023adding} integrates additional features into the decoder of the U-Net architecture to learn task-specific input conditions, such as pose, edges, sketch, and depth etc.\ IP-Adapter~\cite{ye2023ip} propose a unique decoupled cross-attention design for controllable image generation. Differing from these approaches, our method does not rely on reference images; instead, it disentangles the input text prompt into content description and aesthetic description and introduces distinctive improvements to the cross-attention layers.

\section{Methodology}
\label{sec:Methodology}

\subsection{Preliminary}
Since the method proposed in this paper is based on the Stable Diffusion~\cite{rombach2022high} (SD), we first provide a brief overview of it. SD is a latent diffusion model (LDM) capable of transforming Gaussian noise into high-fidelity images through an iterative denoising process. LDM operates diffusion processes in a latent space, requiring an autoencoder that includes an encoder and decoder. We denote the encoder as $\mathcal{E}(\cdot)$, which encodes an image $I$ into the latent space $z=\mathcal{E}(I)$; similarly, the decoder is denoted as $\mathcal{D}(\cdot)$, used for decoding from the latent space back to the image. Given an input text prompt $y$, the text encoder $c_{\theta}(\cdot)$ of a pre-trained CLIP~\cite{radford2021learning} converts it into a text embedding $c_{\theta}(y)$, which will serve as the input condition for LDM. During training, given a latent noise $z_{t}$ at time step $t$ and condition $c_{\theta}(y)$, the denoising network $\epsilon_{\theta}(\cdot)$ aims to predict noise $\epsilon$. This learning process is facilitated by minimizing the following loss function:
\begin{equation}
\mathcal{L}=\mathbb{E}_{z \sim \mathcal{E}(I), y, \epsilon \sim \mathcal{N}(0,1), t}\left[\left\|\epsilon-\epsilon_\theta\left(z_t, t, c_{\theta}(y)\right)\right\|_2^2\right],\label{eq1}
\end{equation}
The denoising network $\epsilon_{\theta}(\cdot)$ is commonly implemented by U-Net~\cite{ronneberger2015u}. When condition $c_{\theta}(y)$ extracted from the text model is integrated into the denoising network $\epsilon_{\theta}(\cdot)$, the cross-attention layer is needed to achieve cross-modal interaction. The process can be described as follows:
\begin{equation}
\mathbf{Q}=\mathbf{W}_{Q} \cdot x, \mathbf{K_c}=\mathbf{W}_{K_c} \cdot c_{\theta}(y), \mathbf{V_c}=\mathbf{W}_{V_c} \cdot c_{\theta}(y),\label{eq2}
\end{equation}
\begin{equation}
\operatorname{Attention}\left(\mathbf{Q}, \mathbf{K_c}, \mathbf{V_c}\right)=\operatorname{softmax}\left(\frac{\mathbf{Q} \mathbf{K_c}^T}{\sqrt{d}}\right)\mathbf{V_c},\label{eq3}
\end{equation}
where $x$ is the spatial feature extracted from latent noise $z$, $\mathbf{W}_{Q}, \mathbf{W}_{K}, \mathbf{W}_{V}$ are learnable projection layers, and $d$ correlates with the number of channels in $x$.

\subsection{The Disentanglement Text Prompts}\label{sec3.2}
This paper aims to further enhance generation quality by integrating aesthetic knowledge across different dimensions. For most fine-tuning approaches~\cite{ruiz2023dreambooth,gal2022image}, the condition is solely derived from the text embedding decoded by the text model from the input text prompt, which encompasses high-level semantic information about the crucial objects and corresponding attributes of an image. In this case, even if there are some aesthetic words in the input text prompt, after several transformer layers, the information can easily \textit{drown} in the process of self-attention with other words, resulting in a minimal cross-modal interaction contribution in U-Net and unsatisfactory performance. On the other hand, the excessive inclusion of aesthetic words, which makes the input prompt overly long, could lead to the inability to generate certain subjects within the prompt.

As illustrated in \cref{Figure 1}, to solve this problem, we initially disentangle the input text prompt of text-to-image synthesis into content and aesthetic input, with aesthetic input $y_{aes}$ being the fine-grained aesthetic labels we introduce, and content input $y$ about the depiction of the main subject and associated attributes in the image. Our starting point comes from the belief that the model can disentangle style (i.e., aesthetics in this case) and content, which is well-documented in~\cite{wu2023uncovering}. 

To enhance the integration of fine-grained aesthetic conditions with the denoising network, we will first introduce the initialization stage of the aesthetic embedding (AesEmb). This phase results in a preprocessed AesEmb that will be consistently utilized throughout both training and inference stages. As shown in \cref{Figure 1}(a), we denote a set of opposing aesthetic labels, where $y_{a}$ denotes a specific aesthetic label (e.g.\ vibrant color, natural lighting, proportional composition, etc.), and $\hat{y}_{a}$ indicates \textit{not having} that label. Notably, we use [identifier] to signify $\hat{y}_{a}$, which is a rare token acting as a unique identifier associated with the aesthetic label (e.g.\ [V], [S])~\cite{ruiz2023dreambooth}. In this context, we employ a rare token to represent $\hat{y}_{a}$ to prevent the semantic prior of the text model from leaking into the negative aesthetic labels. This pair of opposing aesthetic labels $y_p=\{y_{a}, \hat{y}_{a}\}$ is then processed by a frozen CLIP model, yielding a pair of [CLS] tokens, denoted as $t_p=\{t_{cls}, \hat{t}_{cls}\}$.

In practice, more than one set of opposing aesthetic labels is required. Accordingly, we define $N$ sets of aesthetic labels as $\textbf{Y}=\left[y_{p}^{1}, y_{p}^{2}, \ldots, y_{p}^{N}\right]$, where $y_p^i=\{y_{a}^i, \hat{y}_{a}^i\}$ represents the $i$th pair of aesthetic labels. Then we get $N$ sets of [CLS] tokens as $\textbf{T}=\left[t_{p}^{1}, t_{p}^{2}, \ldots, t_{p}^{N}\right]$, where $t_p^i=\{t_{a}^i, \hat{t}_{a}^i\}$ is the $i$th pair of [CLS] tokens generated by the CLIP model. We further concatenate $\textbf{T}$ along the token dimension to obtain our final AesEmb:
\begin{equation}
f_{aes}=\operatorname{concat}\left[t_{p}^{1}, t_{p}^{2}, \ldots, t_{p}^{N}\right] \in \mathbb{R}^{2N \times d},\label{eq5}
\end{equation}
where $f_{aes}$ is the AesEmb and $d$ is the feature dimension. 
It should be emphasized that the initialization of AesEmb requires only a single execution at the start of training and can be cached locally, making the increase in computational cost practically negligible throughout the entire training process.

\begin{figure*}[ht]
	\centerline{\includegraphics[scale=0.706]{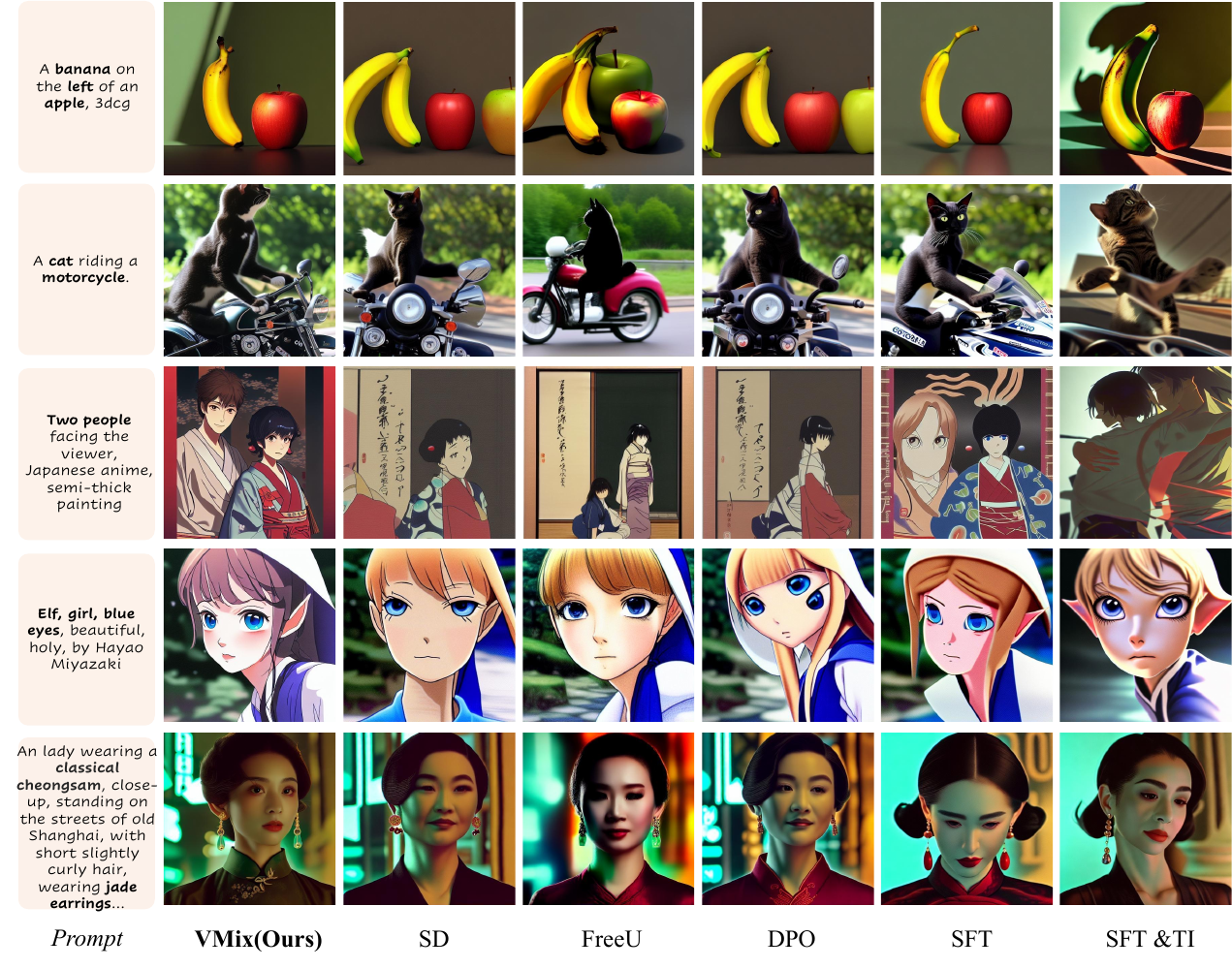}}
	\caption{Qualitative comparison with various state-of-the-art methods. All results are based on Stable Diffusion~\cite{rombach2022high}. Our VMix method outperforms others, significantly enhancing the quality of image generation across various fine-grained aesthetic dimensions.}
	\label{fig4}
\end{figure*}

\begin{figure*}[ht]
	\centerline{\includegraphics[scale=0.71]{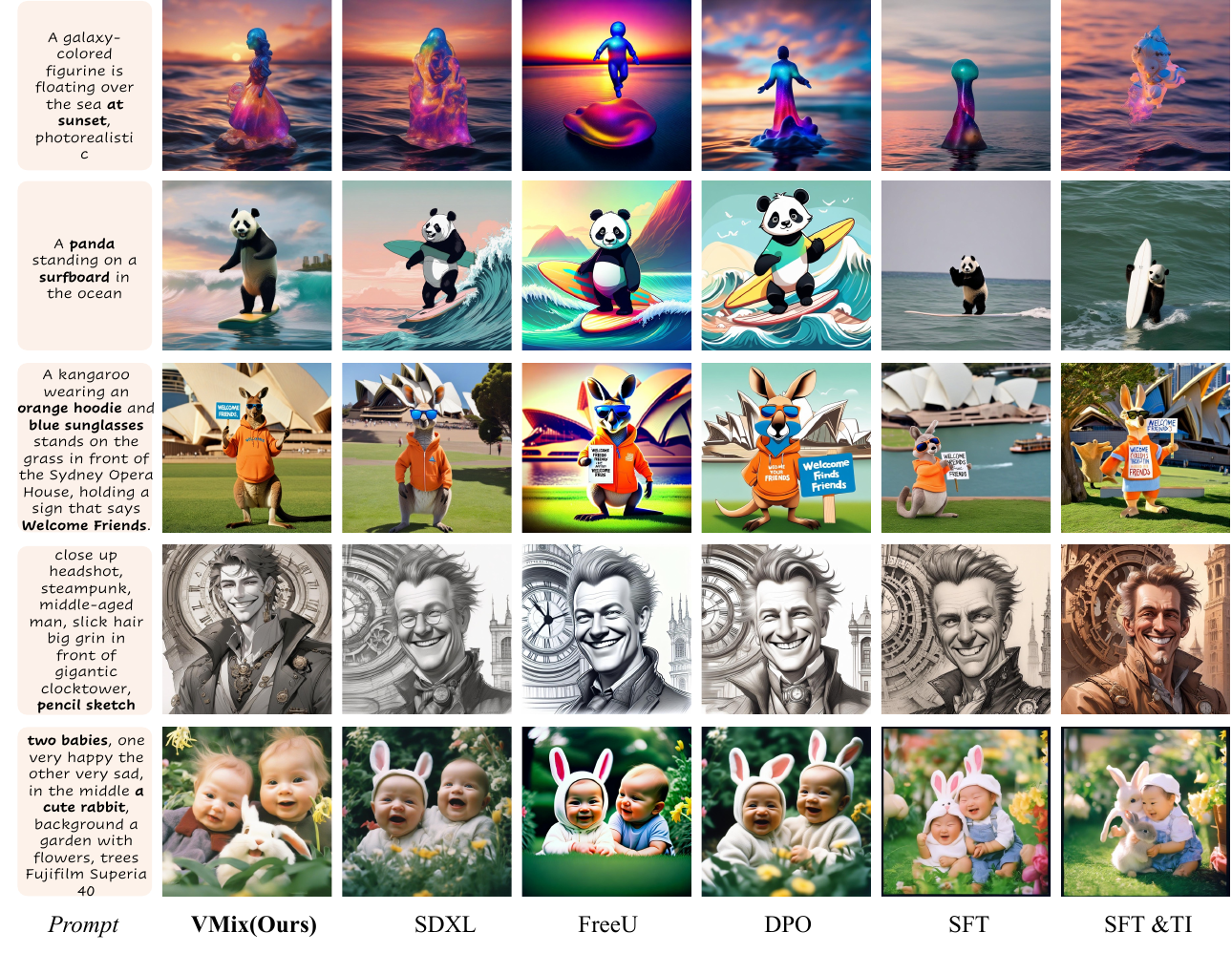}}
	\caption{Qualitative comparison with various state-of-the-art methods. All the results of the methods are based on the SDXL~\cite{podell2023sdxl}. Our VMix method outperforms others, significantly enhancing the quality of image generation.}
	\label{fig5}
\end{figure*}
\subsection{Cross-Attention Mixing Control}
In the previous section, we disentangle the input text prompt into aesthetic input $y_{aes}$ and content input $y$, and introduce the initialization method for AesEmb. In this section, we will further present an effective and nuanced scheme of condition control that leverages fine-grained aesthetic information to enhance the generative quality of the text-to-image model.

\noindent \textbf{Efficient AesEmb Projection Layer.} As is depicted in \cref{Figure 1}(b), we employ the Stable Diffusion~\cite{rombach2022high} (SD) as our text-to-image model, both $y_{aes}$ and $y$ serving as conditions for the model. Similar to the original SD, the $y$ passes through the text model $c_{\theta}(\cdot)$ of CLIP model~\cite{radford2021learning} and is decoded to obtain the text embedding $f_c$, which can be represented by the following equation:
\begin{equation}
f_c=c_{\theta}(y) \in \mathbb{R}^{C \times d},\label{eq6}
\end{equation}
where $C$ is the token length and $d$ is the feature dimension. Due to the inequity of aesthetic labels in the training dataset, where each image is assigned a variable number of aesthetic labels. We consider two approaches to map aesthetic labels into textual features with the same shape as $f_c$. The first method involves directly processing aesthetic labels through the CLIP model to obtain textual feature $c(y_{aes})$, as indicated by \cref{eq6}. Although this approach is straightforward, it introduces certain issues. Firstly, encoding both $y_{aes}$ and $y$ with the text model incurs additional computational costs. More importantly, although we treat different aesthetic dimensions as independent, the attention layers of the text model may potentially compromise this independence.

Given this, we adopt a more efficient method for condition injection. Initially, based on the aesthetic labels included in the input $y_{aes}$, we index the corresponding [CLS] token from AesEmb $f_{aes} \in \mathbb{R}^{2N \times d}$. For the $i$th aesthetic label, we retrieve $t_{a}^i$ if the image has this attribute, otherwise, we obtain $\hat{t}_a^i$. Thus, we can acquire a feature $f_t \in \mathbb{R}^{N \times d}$ reconstituted from $f_{aes}$. Afterward, we use $\mathcal{F}(\cdot)$ to represent the combination of a linear layer and a Layer Normalization~\cite{ba2016layer}, which serve to upscale the token dimension of $f_t$ from $N$ to $C$. To facilitate a gentler condition injection, we also employ a \textit{zero linear} layer, defined as $\mathcal{Z}(\cdot)$, which is a linear layer with both weight and bias initialized to zeros. The entire projection layer is thus computed as follows:
\begin{equation}
f_a=\mathcal{Z}(\mathcal{F}(f_t)),\label{eq7}
\end{equation}
where $f_a$ is the final textual feature projected from aesthetic labels. 
At the start of training, the weights and biases of the zero-initialized linear layer, which serves as the connecting layer, are set to zero. This initialization ensures that fine-tuning the model does not introduce harmful noise, thereby preserving the capabilities of the original pre-trained model~\cite{zhang2023adding}.

\noindent \textbf{Value-Mixed Cross-Attention.} Directly adding aesthetic textual features to the content textual features may compromise rich semantic features, leading to decreased image-text alignment. Since the attention map within the cross-attention layers dictates the probability distribution over the text tokens for each image patch, determining the principal tokens in the image patch~\cite{chefer2023attend}, we aim to preserve this ability inherent in the pre-trained model. This approach ensures a stable enhancement of aesthetic performance while retaining image-text alignment.

To this end, we introduce our value-mixed cross-attention module after each self-attention module in the diffusion U-Net model. We employ a dual-branch cross-attention module, with one branch for feeding in content features $f_c$ and another for aesthetic features $f_a$. The queries for both branches of the cross-attention module are sourced from the spatial feature $x$ of SD, with the keys originating from the content feature $f_c$. However, the sources for the values are distinct; we enable the model to learn a new value for the aesthetic features independently, thus reducing disruption to the original attention map as aesthetic features are fed into the model. The output of cross-attention corresponding to the content description branch shares the same formula as \cref{eq3}, and can be expressed as $\operatorname{Attention}\left(\mathbf{Q}, \mathbf{K_c}, \mathbf{V_c}\right)$. The output of new cross-attention associated with the aesthetic description branch can be formulated as:
\begin{equation}
\mathbf{Q}=\mathbf{W}_{Q} \cdot x, \mathbf{K_c}=\mathbf{W}_{K_c} \cdot f_c, \mathbf{V_a}=\mathbf{W}_{V_a} \cdot f_a,\label{eq8}
\end{equation}
\begin{equation}
\operatorname{Attention}\left(\mathbf{Q}, \mathbf{K_c}, \mathbf{V_a}\right)=\operatorname{softmax}\left(\frac{\mathbf{Q} \mathbf{K_c}^T}{\sqrt{d}}\right)\mathbf{V_a},\label{eq9}
\end{equation}
The cross-attention modules of these two branches share the same attention map $\mathbf{Q} \mathbf{K_c}^T$. Therefore, we only need to add one parameter $\mathbf{W}_{V_a}$ for each cross-attention layer. Subsequently, we add the outputs from the content and aesthetic cross-attention layers to obtain $\hat{x}$, so the complete process of cross-attention mixing control can be represented as follows:
\begin{equation}
\hat{x}=\operatorname{Attention}\left(\mathbf{Q}, \mathbf{K_c}, \mathbf{V_c}\right)+\lambda\operatorname{Attention}\left(\mathbf{Q}, \mathbf{K_c}, \mathbf{V_a}\right),\label{eq10}
\end{equation}
where $\lambda$ is a hyperparameter and set to $1$ during the training phase, $\hat{x}$ is the new spatial feature and will be fed into the subsequent blocks of SD.

\subsection{Training and Inference}\label{sec3.4}
Full-parameter training of models incurs high costs, and while this approach may achieve a higher upper limit of performance, it is inconsistent with our goal of plug-in versatility due to its high degree of customization. For this purpose, during the training phase, we freeze the parameters of the base model, training only the AesEmb Projection Layer and the newly added value in the Value-Mixed Cross-Attention. Additionally, we have incorporated LoRA~\cite{hu2021lora} into some of the model's linear layers and convolutional layers, thereby making the model training process more stable and enhancing the applicability. Upon completion, this segment of the network can be directly extracted to form a plug-and-play module, which can be used to enhancing the aesthetic potential of existing models.

During inference, in addition to the user's prompt $y$, we also require the aesthetic input $y_{aes}$. Unlike the training phase, where the $y_{aes}$ in the training data contains a varying number of positive aesthetic labels (such as "superior light," "inferior color"). During inference, we default to using all positive aesthetic labels as is shown in \cref{Figure 1}(c). This approach aims to enhance the model's generation quality across all aesthetic dimensions. Although we utilized Lora during the training phase, it is not necessary during inference. We will address this aspect in the experimental section with an ablation study.
\section{Experiments}
\label{sec:Experiments}
\subsection{Experiments Setting}
\noindent \textbf{Implementation Details.} We employ AdamW~\cite{loshchilov2017decoupled} optimizer to train our models. The learning rates are set to \(1e-4\) and \(1e-5\) for SD1.5 and SDXL, respectively. The batch size is set to \(256\), and the total number of training steps is \(50,000\) in the experiment. During the inference phase, we employ the DDIM sampler~\cite{song2020denoising} for the sampling process, configuring it with a total of \(25\) timesteps and a classifier-free guidance scale set to \(7.5\), without the use of negative prompts.

\noindent \textbf{Datasets.} As previously discussed, to align the model with high-quality images across various aesthetic dimensions, we finetune our model using a curated dataset of manually selected images. In the dataset construction phase, we prioritize image quality over quantity. Similar to~\cite{dai2023emu}, we initially extracted \(200k\) images from large, publicly available English datasets such as LAION~\cite{schuhmann2021laion}, employing a combination of automatic and human filtering processes. The automatic filtering included aesthetic scoring, OCR scoring, and CLIP scoring. Human filtering was conducted by individuals with a keen aesthetic sense, adhering to universal photography standards to select the finest images. Furthermore, in addition to the content description texts, we annotate these images with categorical labels across different aesthetic dimensions (such as color, lighting, composition, focus) to serve as additional conditions during our training process.

\noindent \textbf{Evaluation Metrics.} We assess our the performance using the MJHQ-30K dataset~\cite{li2024playground} which contains a large number of high-quality, aesthetically pleasing synthetic data. To enhance our evaluation, we created an additional benchmark, LAION-HQ10K, from the LAION~\cite{schuhmann2021laion} collection, including only high-aesthetic and high-resolution real-word images. This set quantifies the gap between our model's generative capabilities and real-world imagery with exceptional aesthetics. For objective evaluation, we use Fréchet Inception Distance (FID), CLIP Scores, and AES Scores\footnote{https://github.com/christophschuhmann/improved-aesthetic-predictor} to measure the overall quality, fidelity to the original prompts, and aesthetic excellence of the generated images.

\subsection{Qualitative Analyses}
\noindent \textbf{Compare to other methods.} To validate the effectiveness of VMix, we compared our model to pre-trained model and systematically conducted further comparisons with state-of-the-art methods such as FreeU~\cite{si2023freeu}, DPO~\cite{wallace2024diffusion}, Textual Inversion(TI)~\cite{gal2022image}, and Supervised Fine-Tuning(SFT). We further apply the well-trained VMix model to personalized models, thereby demonstrating the universality of our approach. It should be noted that, to validate the influence of the training set on the generation results, we proceed to utilize SFT and TI for training. In this configuration, the UNet model will be unfrozen, allowing all parameters to be updated. As depicted in Fig.~\ref{fig4} and Fig.~\ref{fig5}, our VMix significantly outperforms other methods in visual appeal, showcasing remarkable aesthetic performance without compromising the image-text alignment capability. In our comparative analysis with Supervised Fine-Tuning (SFT), it became evident that the model struggles with datasets of exceptionally high quality. This difficulty stems from the presence of complex and abstract samples within the dataset that may surpass the model's current capabilities, potentially leading to a decline in performance. VMix mitigates this challenge by incorporating fine-grained aesthetic supervision signals, which streamlines the learning process for the model and, consequently, enhances the overall performance.

\begin{figure}
    \centering
    \includegraphics[width=0.88\linewidth]{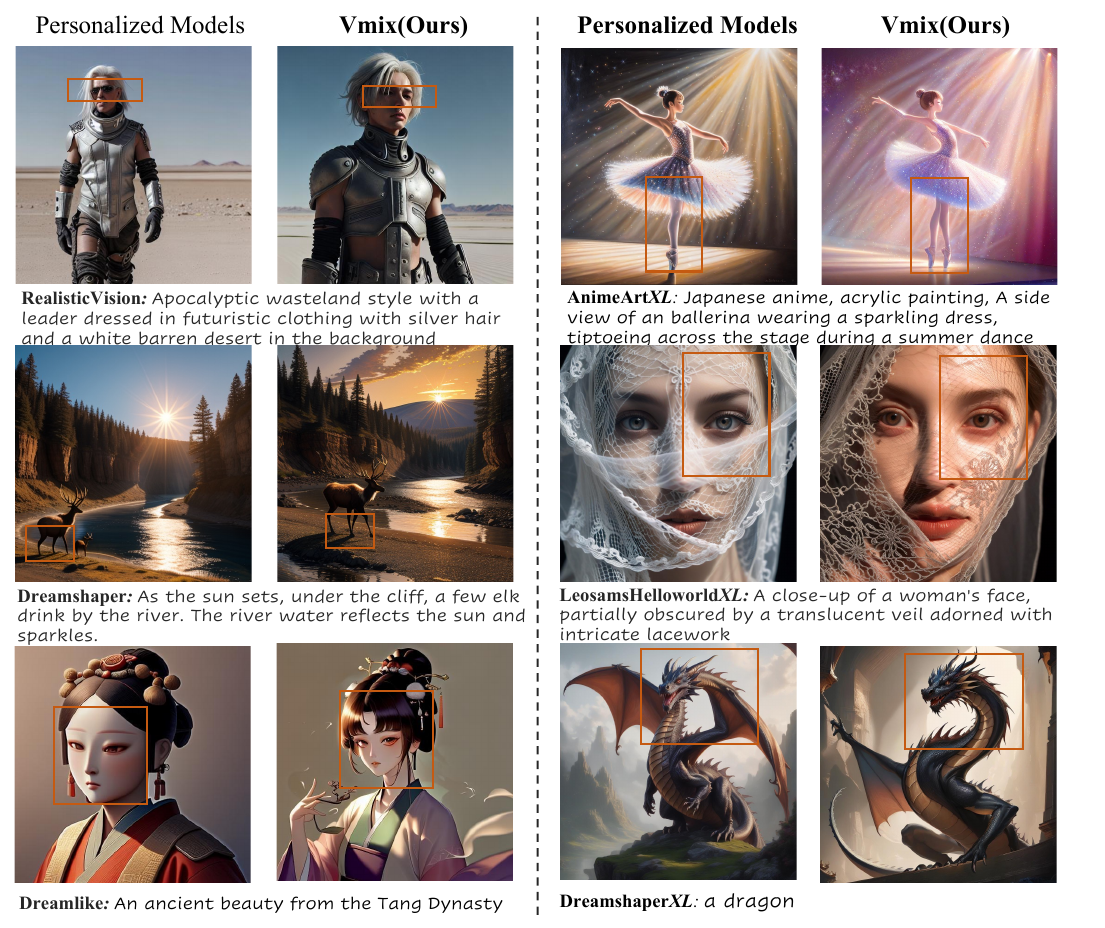}
    \caption{Qualitative results. We compare images generated by VMix-integrated personalized models with those from standard personalized models. On the left are images produced by the personalized model with VMix integration, while on the right are images from the standard personalized model without modifications.}
    \label{fig6}
\end{figure}

\noindent \textbf{Compare to personalized models.} Acting as a versatile plug-and-play adapter, VMix can be directly applied to the personalized model from Civitai\footnote{https://civitai.com}. With the integration of VMix, a notable improvement in the realism and aesthetic appeal of the generated results is expected. See \cref{fig6} for \textit{qualitative results}.

\noindent \textbf{User study.} We further conducted a user study to assess the applicability of VMix as a plug-in. For subjective assessment, 20 evaluators including both aesthetic professionals and non-professionals scored 300 distinct prompts, each yielding 4 generated images. For each case, evaluators need to select the one with the best text fidelity and visual aesthetics from the generation results of the two models. As shown in \cref{fig6}, the results indicate that both pre-trained and open-source models are more favored by users after the application of our VMix method. 

\begin{table}[t]
\centering
\resizebox*{0.425\textwidth}{!}{
\begin{tabular}{lcccc}
\toprule
\textbf{Method} & FID $\downarrow$ & CLIP Score $\uparrow$ & Aes Score $\uparrow$ \\ \hline
SD \cite{rombach2022high} & 28.08 & 30.24 & 5.35 \\
FreeU \cite{si2023freeu} & 27.09 & \textbf{31.00} & 5.36\\
DPO \cite{hu2021lora} &\underline{22.64}  & \underline{30.89} &5.54 \\
Textual Inversion \cite{gal2022image} & 24.72 &  28.92& \underline{5.58}\\
SFT &24.35  & 30.15 & 5.43\\
\hline
\textbf{VMix(Ours)} &\textbf{21.49}  & 30.50 &\textbf{5.79}   \\ \bottomrule
\end{tabular}
}
\caption{Quantitative results on MJHQ-30K benchmark~\cite{li2024playground}. $\uparrow$ stands for higher the better, $\downarrow$ stands for lower the better.}
\label{tab1}
\end{table}

\begin{table}[t]
\centering
\resizebox*{0.425\textwidth}{!}{
\begin{tabular}{lcccc}
\toprule
\textbf{Method} & FID $\downarrow$ & CLIP Score $\uparrow$ & Aes Score $\uparrow$ \\ \hline
SD \cite{rombach2022high} & 25.67 & 32.28 & 5.43 \\
FreeU \cite{si2023freeu} & 28.69 & 32.15 & 5.43\\
DPO \cite{hu2021lora} &\textbf{23.37}  & \underline{32.41} &5.44 \\
Textual Inversion \cite{gal2022image} & 26.62 &  30.97& \underline{5.53}\\
SFT &26.27  & 32.27 & 5.40\\
\hline
\textbf{VMix(Ours)} &\underline{23.92}  & \textbf{32.71} &\textbf{5.68}   \\ \bottomrule
\end{tabular}
}
\caption{Quantitative results on LAION-HQ10K benchmark.}
\label{tab2}
\end{table}

\begin{table}[t]
\centering
\resizebox*{0.425\textwidth}{!}{
\begin{tabular}{lcccc}
\toprule
\textbf{Method} & FID $\downarrow$ & CLIP Score $\uparrow$ & Aes Score $\uparrow$ \\ \hline
Baseline(SD) \cite{rombach2022high} & 28.08 & 30.24 & 5.35 \\ \hline
w/o lora & 21.53 & 30.49 & 5.75\\
w/o vmix &25.64  & 30.16 & 5.52\\
\hline
\textbf{Ours} &\textbf{21.49}  & \textbf{30.50} &\textbf{5.79}   \\ \bottomrule
\end{tabular}
}
\caption{Ablation Study of lora and value-mixed cross-attention. Experiments were conducted on MJHQ-30K benchmark~\cite{li2024playground}.}
\label{tab3}
\end{table}

\begin{figure}
    \centering
    \includegraphics[width=0.85\linewidth]{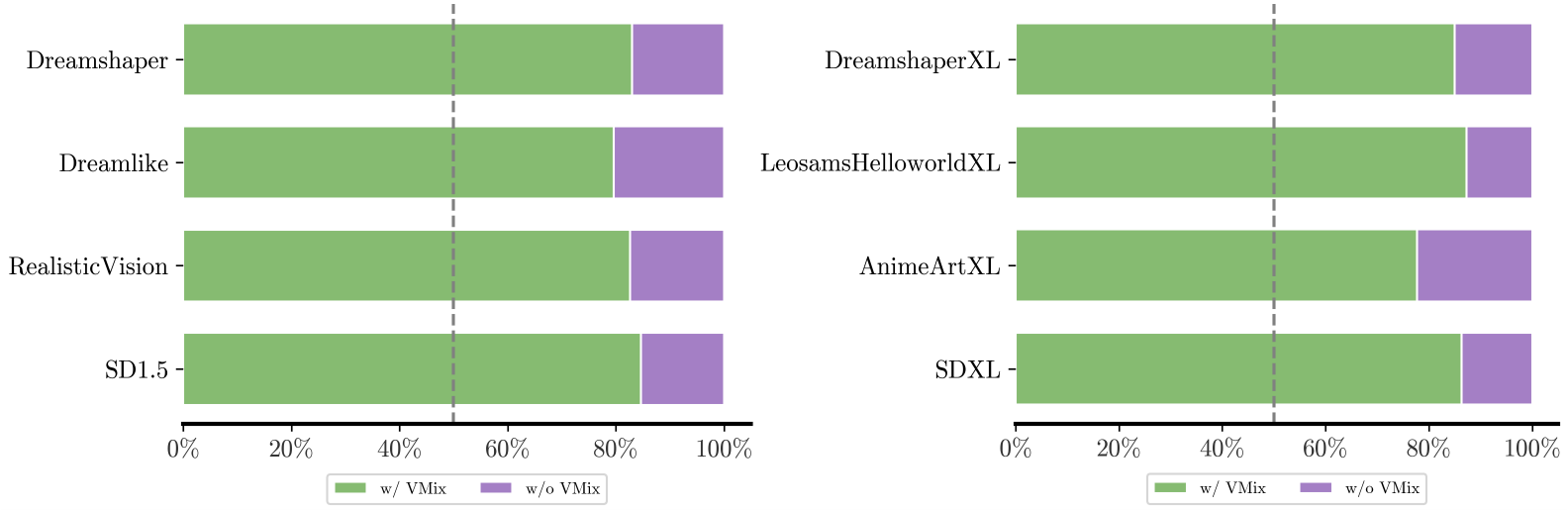}
    \caption{User study. We report the user preference between using VMix and not using VMix.}
    \label{fig7}
\end{figure}

\begin{figure}[ht]
\subfloat[]{
    \begin{minipage}[t]{0.42\linewidth}
        \centering
        \includegraphics[width=0.88\textwidth]{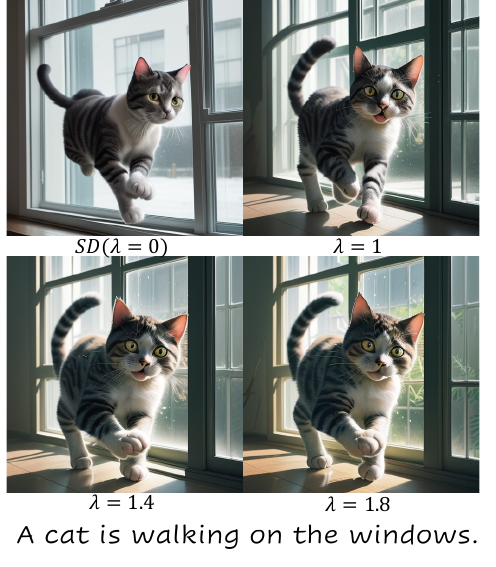}
    \end{minipage}}
\subfloat[]{
    \begin{minipage}[t]{0.54\linewidth}
        \centering
        \includegraphics[width=0.85\textwidth]{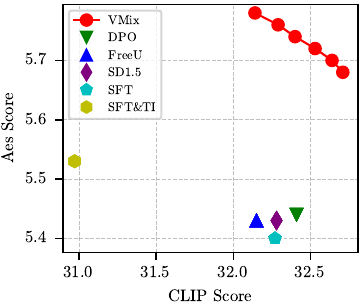}
    \end{minipage}}
    \caption{Ablation Study for $\lambda$ of VMix. (a)Visual performance changes of $\lambda$. (b)Performance metrics for VMix, evaluated across a range of $\lambda$ values from 1 to 2 from right to left. }
    \label{fig8}
\end{figure}


\begin{figure}
    \centering
    \includegraphics[width=0.88\linewidth]{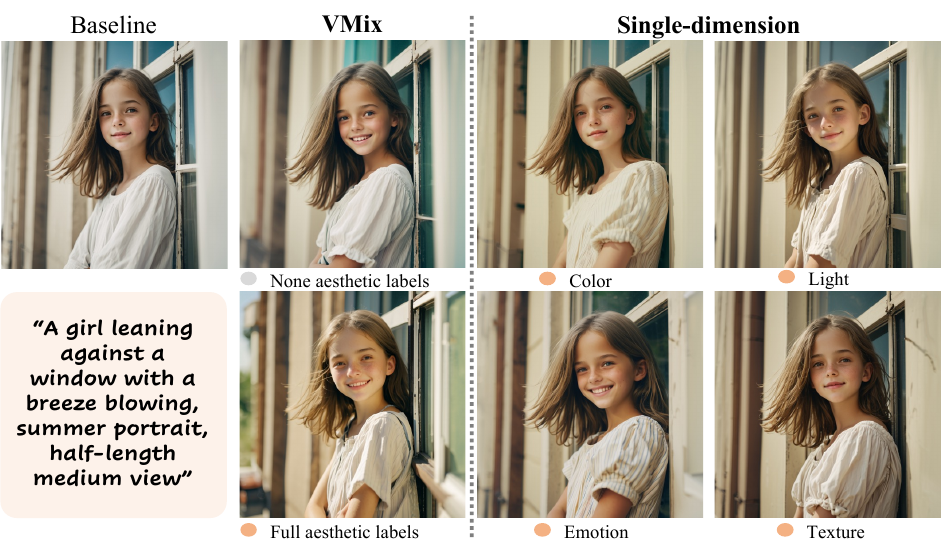}
    \caption{Ablation Study for AesEmb of VMix. \textit{Left}: The effects of using all aesthetic labels versus not using them. \textit{Right}: The effects of using single-dimensional aesthetic labels.}
    \label{fig9}
\end{figure}

\subsection{Quantitative Evaluations}
As shown in \cref{tab1} and \cref{tab2}, we have secured the highest Aes Score on both the MJHQ30K and LAION-HQ10K benchmarks, which strongly demonstrates the significance of VMix in enhancing aesthetics. Our performance in CLIP Score and FID metrics is also commendable, indicating that the incorporation of aesthetic embeddings has not detracted from the model's inherent capabilities. Our findings are consistent with the observations made in Figure \ref{fig6}, where VMix significantly enhances the aesthetic dimensions of images, such as lighting and color. Additionally, imperfections on body parts, such as unrealistic limbs or missing extremities, can be further corrected. Moreover, details in close-up images, like skin texture, are also improved, thereby enhancing the overall aesthetic presentation of the images.

\subsection{Ablation Study}
\noindent \textbf{Effect of $\lambda$.} As illustrated in \cref{fig8}, in Value-Mixed Cross-Attention, $\lambda$ is adjustable during inference. When $\lambda$ increases, the Aes Score is observed to gradually increase, while there is a slight decline in the CLIP
Score. However, our method still maintains a significant advantage over other approaches.

\noindent \textbf{Effect of AesEmb.} As illustrated in \cref{fig9}, we conduct an ablation study on the role of AesEmb. When using only single-dimensional aesthetic label, it can be observed that the image quality improves in specific dimensions. When employing full positive aesthetic labels, the visual performance of the images is superior to the baseline overall. This indicates that the incorporation of AesEmb can enhance the visual appearance of images across various aesthetic dimensions. Throughout this process, we did not utilize LoRA.

\noindent \textbf{Effect of LoRA and VMix cross-attention.} In \cref{tab3}, we examine the impact of lora and value-mixed cross-attention utilized in our method. We discover that both of them can enhance the performance metrics of the baseline. Without value-mixed cross-attention, there is a significant drop in the model's performance, with the Aes Score decreasing from $5.79$ to $5.52$ and the CLIP Score from $30.50$ to $30.16$. This indicates that value-mixed cross-attention plays a significant role in text fidelity and image aesthetics. By combining both, we achieved the best performance.

\section{Conclusion}
\label{sec:Conclusion}
\label{sec:conclusion}
In this paper, we present VMix, which uses the disentangled aesthetic description as an additional condition and employs a cross-attention control method to enhance the performance of model across various aesthetic dimensions. We discover one of the most crucial factors for aligning the model with human expectations is training with decoupled, fine-grained aesthetic labels using a suitable conditional control method. Inspired by these, we proposed an effective conditional control method that significantly improves the generative quality of the model. Extensive experiments validate that VMix surpasses other state-of-the-art methods in terms of text fidelity and visual aesthetics. As a plug-and-play plugin, VMix can seamlessly integrate with open-source models, enhancing aesthetic performance and thereby further promoting the development of the community.
{
    \small
    \bibliographystyle{ieeenat_fullname}
    \bibliography{main}
}

\clearpage
\setcounter{page}{1}

\section{Supplementary}
\label{sec:Supplementary}

\subsection{More Qualitative Comparison}
\label{More Qualitative Comparison}
VMix decouples aesthetic knowledge from content knowledge and introduces a novel conditional control method. To further verify its effectiveness, we provide additional experimental results here.

\begin{figure}[ht]
\centering
\includegraphics[scale=0.4]{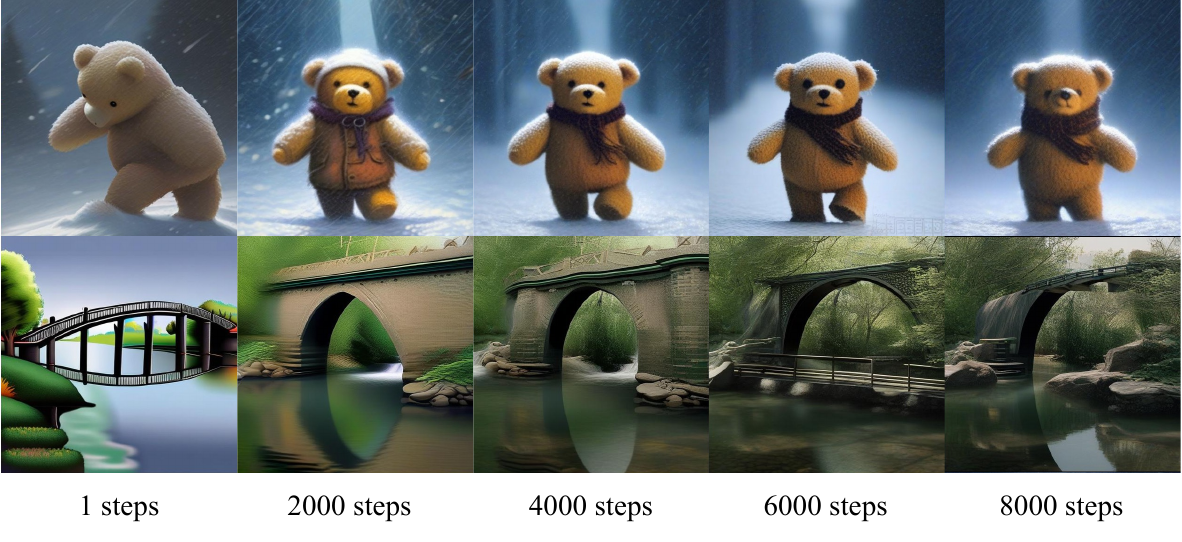}
    \caption{Visualization results of different training steps. Prompts: (1) A teddy bear walking in the snowstorm. (2) A bridge is depicted in the water.}
    \label{Figure 100}
\end{figure}

\noindent \textbf{Training Stability.} In \cref{sec:Methodology}, we introduced Value-Mixed Cross-Attention(vmix cross-attention), which learns a single value for the projected aesthetic embedding. This approach might seem counterintuitive, particularly since in the aesthetic branch, the Q (Query) and V (Value) originate from different sources. In practice, AesEmb is initialized from the [CLS] tokens of the text model, ensuring semantic alignment with the original text embedding. Furthermore, the projected aesthetic embedding must pass through a zero-initialized linear layer before it can finally enter the vmix cross-attention. This process further ensures a gentle injection of aesthetic knowledge, minimizing disruption to the original model. As shown in \cref{Figure 100}, the entire training process is relatively stable, with gradual improvements in lighting, color, and other visual aspects.

\begin{figure}[ht]
\centering
\includegraphics[scale=0.33]{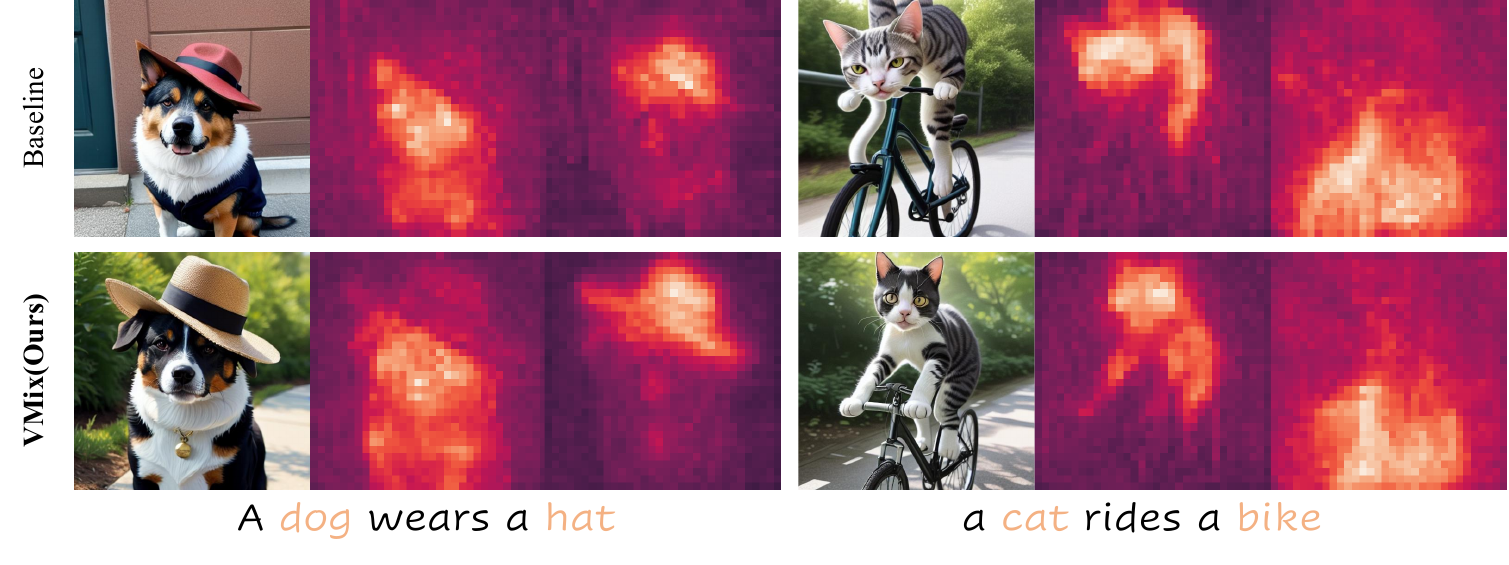}
    \caption{Visualization results of attention maps. VMix is capable of maintaining attention maps that closely resembles that of the baseline(SD~\cite{rombach2022high}) while further enhancing the quality of the generated images.}
    \label{Figure 101}
\end{figure}

\noindent \textbf{Effect of VMix Cross-Attention.} In vmix cross-attention, we designed the aesthetic branch and the content branch to share the same attention map, denoted as $\mathbf{Q} \mathbf{K_c}^T$, to prevent the injection of aesthetic knowledge from significantly impairing the model's text fidelity capabilities. As demonstrated in \cref{Figure 101}, VMix produces an attention score map that closely resembles that of the baseline. After the denoising process, we can obtain a generated image that maintains a layout roughly equivalent to the baseline but with enhanced quality. This indicates that our vmix cross-attention allows the model to concentrate more on refining overall details, thereby directly boosting the model's performance across various fine-grained aesthetic dimensions, including lighting, color, and more.

\begin{figure}[ht]
\centering
\includegraphics[scale=0.58]{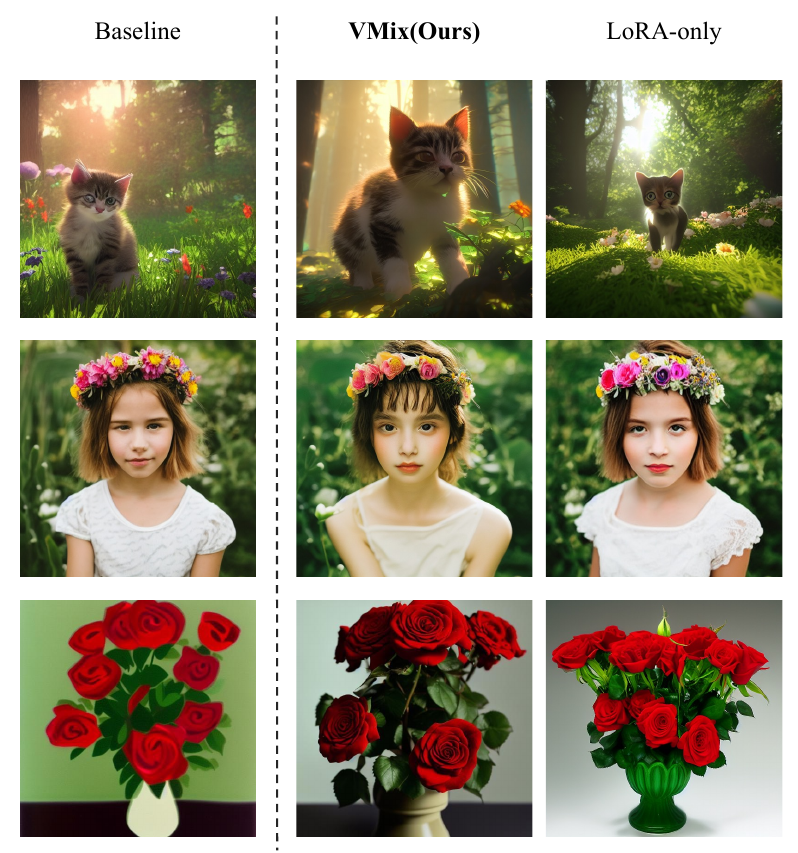}
    \caption{Qualitative comparison. Prompts: (1) Kitten in the forest with flowers with sunlight on them, Cinematic lighting, Unreal Engine 5. (2) Close-up of a young girl wearing a flower crown in the garden, portrait. (3) A green vase with several red roses in it.}
    \label{Figure 102}
\end{figure}

\noindent \textbf{Comparison with LoRA.} Although in \cref{sec:Experiments}, we compared our method with SFT and textual inversion on the same dataset, the use of LoRA in our training process might obscure the enhancement of the final results. To clarify this, we trained a model with only LoRA on the same dataset. As shown in \cref{Figure 102}, our method significantly improves upon the SD~\cite{rombach2022high} more than the approach that uses only LoRA for training.

\subsection{More Visualization}
\begin{figure}[ht]
\centering
\includegraphics[scale=0.58]{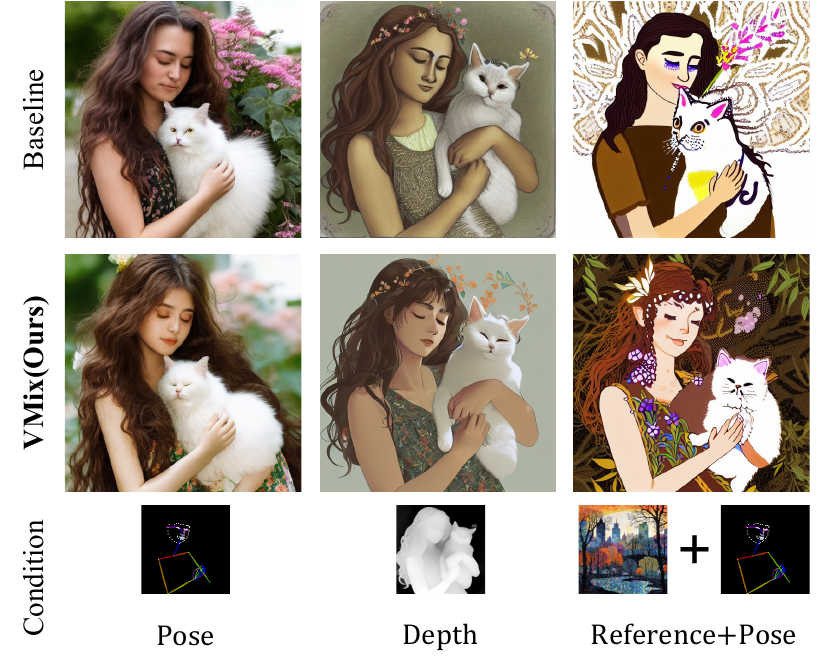}
    \caption{Qualitative results about VMix with ControlNet\cite{zhang2023adding} and IP-Adapter~\cite{ye2023ip}. Prompt: a young woman with long, wavy brown hair. she is wearing a sleeveless floral dress with a pattern of various flowers and leaves. the woman is holding a white, fluffy cat close to her face, seemingly in a moment of affection and joy. her eyes are closed, suggesting she is savoring the moment.}
    \label{Figure 104}
\end{figure}

\begin{figure*}[t]
\centering
\includegraphics[scale=0.995]{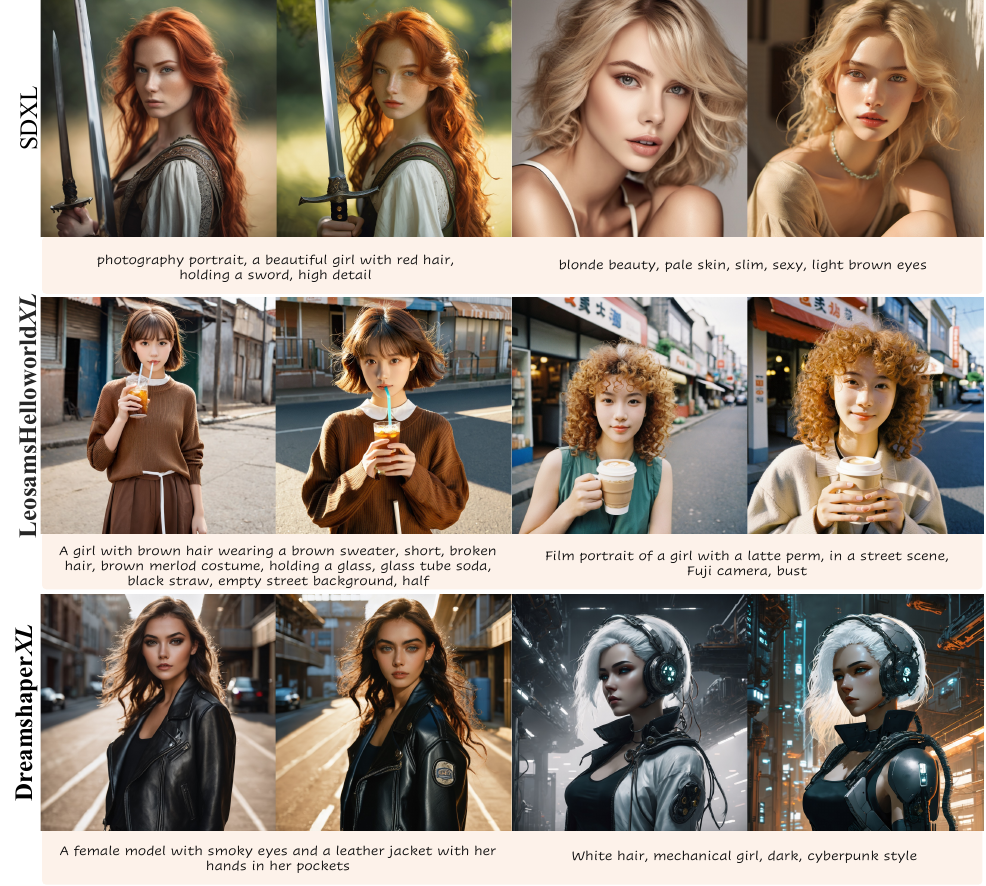}
    \caption{Qualitative comparison between results with VMix(on the right) and without VMix(on the left), shows that VMix significantly enhances the quality of image generation.}
    \label{Figure 105}
\end{figure*}

We apply VMix with ControlNet\cite{zhang2023adding} and IP-Adapter~\cite{ye2023ip}. As shown in \cref{Figure 104}, VMix can be compatible with these standard methods and generates images with better visual aesthetics. As shown in the \cref{Figure 105}, we have provided additional comparative results with SDXL~\cite{podell2023sdxl} as well as its variants. When VMix is incorporated, the generated results show significant improvements across various aesthetic dimensions, offering enhanced visual performance.

\begin{figure*}[t]
\centering
\includegraphics[scale=0.995]{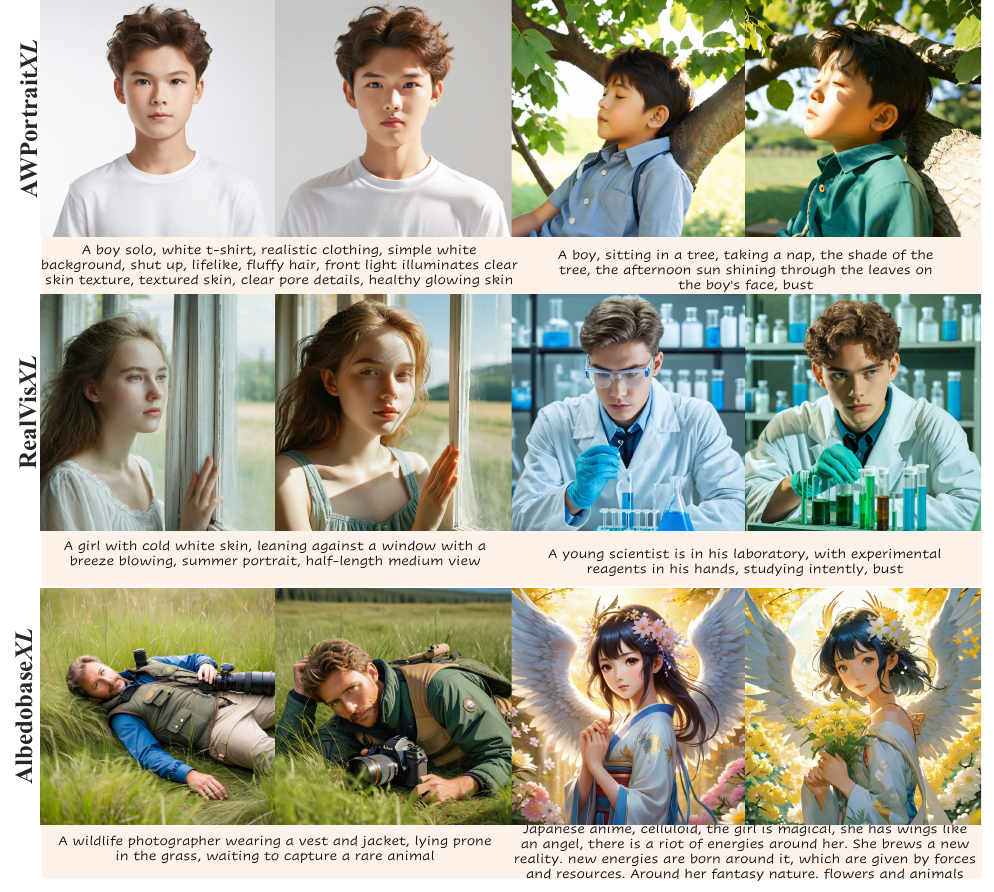}
    \caption{Qualitative comparison between results with VMix(on the right) and without VMix(on the left), shows that VMix significantly enhances the quality of image generation.}
    \label{Figure 106}
\end{figure*}

\subsection{Limitations}
\label{Limitations}
Despite the superior aesthetic generation effects achieved by VMix, it still has several limitations: (1) Currently, the aesthetic labels form a closed set, and the included aesthetic dimensions may not cover all necessary aspects. Although we have confirmed the effectiveness of our current method, VMix's performance is inevitably impacted. We intend to further optimize this aspect in our future work. (2) Images generated by VMix may exhibit a bias towards certain specific objects. For instance, when we attempt to generate concrete objects found in real life, such as cups or mobile phones, and include all aesthetic labels, including emotional ones, during the inference phase, the resulting images might unexpectedly depict humans. This is because, in the training set, emotional labels are typically associated only with people or animals. Consequently, these labels may become bound to specific entities during the training phase, potentially affecting the outcomes of the inference process.




\end{document}